\pdfoutput=1

\documentclass[11pt]{article}

\usepackage[]{naacl2021}

\usepackage{times}
\usepackage{latexsym}

\usepackage[T1]{fontenc}

\usepackage[utf8]{inputenc}

\usepackage{microtype}

\usepackage{graphicx}
\usepackage{url}
\usepackage{amsmath}
\usepackage{breqn}
\usepackage{makecell}
\usepackage{tabularx}
\usepackage{bm}

\newcommand*{\email}[1]{\texttt{#1}}
\newcommand*{\affaddr}[1]{#1}
\newcommand*{\affmark}[1][*]{\textsuperscript{#1}}

\usepackage{microtype}
\usepackage{amsmath}

\newcommand{\ZeroShotAllDatasets}{
\begin{table}[t!]
\centering
\resizebox{\columnwidth}{!}{\begin{tabular}{c | c c c | c c c c } \Xhline{2\arrayrulewidth}
             \textbf{Target Dataset} &  \multicolumn{3}{c|}{\textbf{WikiTransfer}} & \multicolumn{4}{c}{\textbf{Other Transfer}}   \\ \Xhline{\arrayrulewidth}
              CNNDM & \textbf{39.11} & \textbf{17.25} & \textbf{35.73} & 36.81 & 14.18 & 32.62 &  (Reddit)   \\
              XSum & \textbf{31.85} & \textbf{10.44} & \textbf{23.75} & 24.04 & 6.43 & 18.99 &  (Reddit)   \\ 
             Reddit & \textbf{21.47} & 4.10 & 17.62 & 21.37 & \textbf{4.14} & \textbf{17.76} &  (CNNDM)  \\
             BigPatent & \textbf{35.58} & \textbf{10.91} & \textbf{31.53} & 33.57 & 9.34 & 25.76 & (CNNDM)    \\
\Xhline{2\arrayrulewidth}
\end{tabular}}
\caption{Comparison of ROUGE-1/2/L zero-shot transfer performance from dataset-specific WikiTransfer vs. transfer from another dataset. The dataset from which zero-shot transfer performed the best is in parentheses.}
\label{tab:zero_shot}
\end{table}
}

\newcommand{\UnsupComparison}{
\begin{table}[t!]
\tiny
\centering
\resizebox{.8\columnwidth}{!}{\begin{tabular}{c | c c c} \Xhline{2\arrayrulewidth}
             \textbf{Model} & \multicolumn{3}{c}{\textbf{ROUGE-1/2/L}}   \\
              WikiTransfer & \textbf{39.11} & \textbf{17.25} & \textbf{35.73} \\
              TED \cite{yang-etal-2020-ted}  & 38.73 & 16.84 & 35.40  \\
             \Xhline{2\arrayrulewidth}
\end{tabular}}
\caption{A comparison of our approach to the unsupervised pretraining of TED \cite{yang-etal-2020-ted}, showing the superior performance and generalizability of our approach versus the TED model, which focused specifically on the news domain.}
\label{tab:unsup_comparison}
\end{table}
}

\newcommand{\SentenceChoiceComparison}{
\begin{table}[t!]
\centering
\resizebox{\columnwidth}{!}{\begin{tabular}{c | c c c | c c c | c c c} \Xhline{2\arrayrulewidth}
            \textbf{Target Dataset} & \multicolumn{3}{c|}{\textbf{First M Sents}} & \multicolumn{3}{c|}{\textbf{IND-ORIG}} & \multicolumn{3}{c|}{\textbf{IND-ORIG-P}}   \\  \Xhline{2\arrayrulewidth}
             CNNDM &  40.14 & 17.71 & 36.66  & 37.62 & 15.15 & 34.21   &  37.85 & 15.32 & 34.39 \\
             XSum & 31.80 & 10.46 & 23.66 & 29.95 & 9.37 & 21.78  & 30.22 & 9.79 & 23.23 \\
             \Xhline{2\arrayrulewidth}
\end{tabular}}
\caption{A comparison of the effect of summary sentence choice for WikiTransfer on zero-shot transfer ROUGE-1/2/L performance.}
\label{tab:sentence_choice_comparison}
\end{table}
}
\newcommand{\DatasetSizeComparison}{
\begin{table}[t!]
\centering
\resizebox{\columnwidth}{!}{\begin{tabular}{c | c c c | c c c  } \Xhline{2\arrayrulewidth}
            \textbf{Intermediate Dataset Size} &  \multicolumn{3}{c|}{\textbf{CNNDM}} &  \multicolumn{3}{c}{\textbf{XSum}}  \\  \Xhline{2\arrayrulewidth}
              10k & 39.48 & 17.79 & 36.3 & 21.59 & 4.85 & 16.28 \\
              100k & 39.92 & 17.65 & 36.5 & 31.52 & 10.86 & 23.94 \\
              250k & 40.10 & 17.70 & 36.62 & 31.39 & 10.27 & 23.43 \\
              400k & 40.14 & 17.71 & 36.66 & 31.80 & 10.46 & 23.66 \\
             \Xhline{2\arrayrulewidth}
\end{tabular}}
\caption{A comparison of the effect of dataset size of the unsupervised intermediate fine-tuning data on the zero-shot transfer ROUGE-1/2/L performance.} 
\label{tab:dataset_size_comparison}
\end{table}
}

\newcommand{\AblationsLRBinM}{
\begin{table}[t!]
\centering
\resizebox{\columnwidth}{!}{\begin{tabular}{c | c c c | c c c } \Xhline{2\arrayrulewidth}
            \textbf{Ablation} & \multicolumn{3}{c|}{ \textbf{CNNDM}} & \multicolumn{3}{c}{\textbf{XSum}}  \\ \hline
              LR=3e-6 & 40.14 & 17.71 & 36.66 & 27.60 & 8.62 & 20.93 \\
              LR=3e-5 & 39.73 & 16.94 & 36.24 & 31.80 & 10.46 & 23.66 \\ \hline
              LR=3e-6, No-bin & 39.11 & 16.98 & 35.66 & 22.78 & 5.66 & 17.16  \\ \hline
              LR=3e-6, bin, M=1 & 37.45 & 14.72 & 32.52 & 27.60 & 8.62 & 20.93   \\ 
              LR=3e-6, bin, M=3  &  40.14 & 17.71 & 36.66 & 27.98 & 9.59 & 23.11  \\ \hline
             \Xhline{2\arrayrulewidth}
\end{tabular}}
\caption{Ablation studies on the effect of learning rate, the use of extractive bin for data filtering and the choice of M in intermediate fine-tuning on ROUGE-1/2/L performance on CNNDM and XSum validation sets. 
}
\label{tab:ablations_lr_bin_m}
\end{table}
}

\newcommand{\TableAll}{
\begin{table}[t!]
\tiny
\centering
\resizebox{\columnwidth}{!}{\begin{tabular}{ c | c c c | c c c | c c c } \Xhline{2\arrayrulewidth}
             \textbf{Target Dataset} &  \multicolumn{9}{c}{CNNDM}  \\
             \Xhline{2\arrayrulewidth}
             \textbf{Transfer from} & \multicolumn{3}{c|}{WikiTransfer} & \multicolumn{3}{c|}{Reddit} & \multicolumn{3}{c}{BART} \\ \Xhline{2\arrayrulewidth}
            \textbf{0} & \textbf{39.11} &  \textbf{17.25} & \textbf{35.73} & 36.81 & 14.18 & 32.62 & 35.98 & 15.10 & 32.97  \\  \Xhline{2\arrayrulewidth}
            \textbf{10} & \textbf{39.10} & \textbf{16.98} & \textbf{35.84} & 38.26 & 16.34 & 34.76 &  38.55 & 16.56 & 34.97 \\
             \textbf{10-a} & 39.39 & 16.92 & 36.00 & 39.12 & 16.90 & 35.44 & \textbf{39.78} & \textbf{17.11} & \textbf{36.38} \\
              \textbf{10-c} & \textbf{39.16} & \textbf{16.96} & \textbf{35.92} & 38.99 & 16.83 & 35.43 &  38.98 & 16.68 & 35.41 \\
             \Xhline{2\arrayrulewidth}
             \textbf{100} &  \textbf{40.55} & \textbf{18.01} & \textbf{37.03} & 40.13 & 17.88 & 36.67 & 40.14 & 17.88 & 36.62 \\
             \textbf{100-a} & \textbf{42.08} &  \textbf{18.93} & \textbf{38.83} & 40.94 & 18.52 & 37.00 & 40.47 & 18.18 & 37.07  \\
             \textbf{100-c} & 41.12 & 18.34 & 37.51  & 40.84 & 18.09 & 37.28 & 
             \textbf{41.36} & \textbf{18.59} & \textbf{37.77} \\
             \Xhline{2\arrayrulewidth}
             \multicolumn{4}{c}{}   \\
            \Xhline{2\arrayrulewidth} 
            \textbf{Target Dataset} &   \multicolumn{9}{c}{XSum} \\
            \Xhline{2\arrayrulewidth}
            \textbf{Transfer from} & \multicolumn{3}{c|}{WikiTransfer} & \multicolumn{3}{c|}{Reddit} & \multicolumn{3}{c}{BART} \\
            \Xhline{2\arrayrulewidth}
            \textbf{0} & \textbf{31.85} & \textbf{10.44} & \textbf{23.75} & 24.04 & 6.43 & 18.99 & 19.87 & 2.75 & 15.66 \\
            \Xhline{2\arrayrulewidth}
            \textbf{10} & \textbf{34.95} & \textbf{12.61} & \textbf{26.58} & 30.69 & 10.22 & 23.29 & 22.45 & 5.94 & 17.23  \\
            \textbf{10-a} & \textbf{34.98} & \textbf{12.73}& \textbf{26.79} & 31.03 & 10.23 & 23.29 & 26.10 & 8.19 & 20.18 \\
             \textbf{10-c}  & \textbf{35.17} & \textbf{12.76} & \textbf{26.80} & 31.25 & 10.54 & 23.73  & 28.28 & 9.13 & 21.61 \\
            \Xhline{2\arrayrulewidth}
             \textbf{100} & \textbf{36.92}& \textbf{14.09}& \textbf{28.44} & 34.17 & 12.64 & 26.37 & 35.17 & 13.29 & 27.20  \\
             \textbf{100-a} & \textbf{36.87}& \textbf{14.18} & \textbf{28.62} & 31.75 & 11.12 & 24.49 & 28.85 & 9.46 & 22.28 \\
            \textbf{100-c} & \textbf{37.26} & \textbf{14.20} & \textbf{28.85} & 36.14 & 13.65 & 27.97 & 36.65 & 14.05 & 28.57 \\
            \Xhline{2\arrayrulewidth}
            \multicolumn{4}{c}{}   \\
            \Xhline{2\arrayrulewidth}
            \textbf{Target Dataset} &  \multicolumn{9}{c}{Reddit} \\
            \Xhline{2\arrayrulewidth}
            \textbf{Transfer from} & \multicolumn{3}{c|}{WikiTransfer} & \multicolumn{3}{c|}{CNNDM} & \multicolumn{3}{c}{BART} \\
            \Xhline{2\arrayrulewidth}
            \textbf{0}  & \textbf{21.47} & 4.10 & 17.62 & 21.37 & \textbf{4.14}& \textbf{17.76} & 18.66 & 2.90 & 15.33 \\
            \Xhline{2\arrayrulewidth}
            \textbf{10}  & \textbf{27.88}& \textbf{7.62}& \textbf{22.09} & 26.55 & 6.83 & 21.29 & 19.37 & 3.51 & 15.72 \\
            \textbf{10-a} & \textbf{28.07}& \textbf{7.70}& \textbf{22.47} & 26.88 & 6.95 & 21.46 & 21.39 & 4.57 & 17.22 \\
             \textbf{10-c} & \textbf{28.42} & \textbf{7.88} & \textbf{22.32} & 27.20 & 7.14 & 21.67 & 20.42 & 3.97 & 16.45 \\
            \Xhline{2\arrayrulewidth}
              \textbf{100}  & \textbf{29.87} & \textbf{8.93} & \textbf{23.31} & 28.90 & 8.42 & 22.56 & 29.66 & 8.88 & 23.12 \\
             \textbf{100-a} & \textbf{30.54}& \textbf{9.24}& \textbf{24.31} & 29.28 & 8.51 & 23.28 & 28.96 & 8.39 & 22.80  \\
             \textbf{100-c} & 30.56 & 9.22 & \textbf{24.38} & \textbf{30.78} & \textbf{9.45} & 24.14 &\textbf{30.78} & 9.22 & 23.32   \\
            \Xhline{2\arrayrulewidth}
             \multicolumn{4}{c}{}   \\
            \Xhline{2\arrayrulewidth}
            \textbf{Target Dataset} &  \multicolumn{9}{c}{BigPatent} \\
            \Xhline{2\arrayrulewidth}
          \textbf{Transfer from} & \multicolumn{3}{c|}{WikiTransfer} & \multicolumn{3}{c|}{CNNDM} & \multicolumn{3}{c}{BART} \\
          \Xhline{2\arrayrulewidth}
          \textbf{0}   &  \textbf{35.58}& \textbf{10.91}& \textbf{31.53} & 33.57 & 9.34 & 25.76 & 32.56 & 9.64 & 29.27 \\
          \Xhline{2\arrayrulewidth}
              \textbf{10}  & \textbf{37.06}& \textbf{11.58}& \textbf{32.37} & 35.76 & 10.62 & 30.63 & 34.48 & 10.76 & 30.56 \\  
              \textbf{10-a} & \textbf{37.73} & \textbf{12.40} & \textbf{32.89} & 36.83 & 11.33 & 30.95 &  36.11 & 11.40 & 32.04 \\  
              \textbf{10-c}  & \textbf{37.64} & \textbf{12.24} & \textbf{33.05} &  36.11 & 10.84 & 30.64  & 33.99 & 10.48 & 30.45 \\  \Xhline{2\arrayrulewidth}
              \textbf{100}  & \textbf{39.61} & \textbf{13.53} & 33.86 & 39.35 & 13.03 & \textbf{33.88} & 39.06 & 13.04 & 33.61 \\ 
                \textbf{100-a}  & \textbf{40.95} & \textbf{ 14.05 } & \textbf{35.03} & 38.88 & 12.69 & 32.88 & 38.77 & 12.88 & 33.55 \\  
                \textbf{100-c} & \textbf{39.87} & \textbf{13.76} & 34.32 & 39.74 & 13.45 & \textbf{34.49} & 39.46 & 13.37 & 34.28 \\  \Xhline{2\arrayrulewidth}
\end{tabular}}
\caption{A comparison of transfer results across datasets, training dataset size, data augmentation techniques, showing the generalizable and robust performance of our models transferred from WikiTransfer.}
\label{tab:table_all}
\end{table}
}

\newcommand{\TableComparePegasus}{
\begin{table*}[ht!]
\centering
\resizebox{2\columnwidth}{!}{\begin{tabular}{c | c c c | c c c | c c c | c c c | c c c | c c c | c c c } \Xhline{2\arrayrulewidth}
            \textbf{Target Dataset} &  \multicolumn{9}{c|}{WikiTransfer} &  \multicolumn{9}{c}{Pegasus \cite{zhang2019pegasus}}  \\
          \Xhline{2\arrayrulewidth} 
          \# training samples   &  \multicolumn{3}{c|}{0} & \multicolumn{3}{c|}{10} & \multicolumn{3}{c|}{100} & \multicolumn{3}{c|}{0} & \multicolumn{3}{c|}{10}  & \multicolumn{3}{c}{100} \\
          \Xhline{2\arrayrulewidth} 
          CNNDM   &  \textbf{39.11} & \textbf{17.25} & \textbf{35.73} & \textbf{39.39} & \textbf{16.92} & \textbf{36.00} & \textbf{42.08} & \textbf{18.93} & \textbf{38.83} & 32.90 & 13.28 & 29.38 & 37.25 & 15.84 & 33.49  & 40.28 & 18.21 & 37.03 \\
          XSum   &  \textbf{31.85} & \textbf{10.44} & \textbf{23.75} & \textbf{35.17} &  \textbf{12.76} & \textbf{26.80} & 37.26 & 14.20 & 28.85 & 19.27 & 3.00 & 12.72 & 19.39 & 3.45 & 14.02  & \textbf{39.07} &  \textbf{16.44} &  \textbf{31.27} \\
          Reddit   &  \textbf{21.47} & \textbf{4.10} &  \textbf{17.62} & \textbf{28.42} &  \textbf{7.88} & \textbf{22.32} & \textbf{30.56 } &  \textbf{9.22} &  \textbf{24.38} & 14.66 & 3.06 & 10.17 & 15.36 & 2.91 & 10.76  & 16.64 & 4.09 & 12.92 \\
          BigPatent   &  \textbf{35.58} &  \textbf{10.91} &  \textbf{31.53} & \textbf{37.73}  & \textbf{12.40} &  \textbf{32.89} & \textbf{40.95} &  \textbf{14.05} & \textbf{35.03} & 25.61& 6.56 & 17.42 & 28.87 & 8.30 & 19.71  & 33.52 & 10.82 & 22.87 \\
            \Xhline{2\arrayrulewidth}
\end{tabular}}
\caption{A comparison of zero and few-shot performance between our best-performing WikiTransfer model (-a in the case of CNNDM and BigPatent and -c for XSum and Reddit) and the zero and few-shot results reported in \citet{zhang2019pegasus}.}
\label{tab:table_compare_pegasus}
\end{table*}
}

\newcommand{\TableHumanEvaluation}{
\begin{table}[t!]
\centering
\resizebox{\columnwidth}{!}{\begin{tabular}{c | c c |  c c} \Xhline{2\arrayrulewidth}
            \textbf{Target Dataset} &  \multicolumn{2}{c|}{CNNDM} &  \multicolumn{2}{c}{XSum}  \\
              &  Relevance & Consistency & Relevance & Consistency \\
          \Xhline{2\arrayrulewidth} 
          0   &  \textbf{4.37} & \textbf{4.71} &  3.75* &  \textbf{3.75} \\
          10-a   &  4.31 & 4.76 &  3.77* & 4.10 \\
          100-a   &  4.25 & 4.86 & \textbf{4.00}  &  4.04 \\
          Full supervision   &   4.31 & 4.86 &  4.11  & 3.98 \\
            \Xhline{2\arrayrulewidth}
\end{tabular}}
\caption{Summary relevance and factual consistency across CNNDM and XSum datasets with varying amounts of training data. All results except those with an asterisks do not differ in a statistically significant way (p-value of 0.05) from the full supervision score. Bold results emphasize the least amount of data to achieve statistically indistinguishable results from the fully-supervised results.}
\label{tab:table_human_evaluation}
\end{table}
}

\newcommand{\ExampleSummaries}{
\begin{table}[t!]
\centering
\small
\begin{tabularx}{\columnwidth}{|X|}
\hline
\textbf{Source Document:} \textcolor{blue}{Ms Jones} told BBC Radio Wales she did not want to give up being an AM to go to Brussels to replace Nathan Gill, UKIP Wales leader. Mr Gill has been told by the UKIP assembly group and the UKIP party chairman Steve Crowther to stop "double-jobbing" as an AM and MEP. Mr Gill said those making such calls were doing it out of "malice". "We've got Brexit now and I think that, possibly, it may be best to leave that role unfilled," Ms Jones told the Good Morning Wales programme. "I'm surprised I've not been formally asked what I'd like to do." Ms Jones, the South Wales West AM, is one of two people who could take up the role of UKIP Wales MEP if Mr Gill made it vacant - the other being South Wales East AM David Rowlands...
 \\ \hline \hline

\textbf{0:} \textcolor{red}{Lorraine} Jones is a Welsh Labour Party Member of the Welsh Assembly for South Wales West.
 \\ \hline
\textbf{10-a:} \textcolor{red}{Lorraine} Jones is a Welsh Labour member of the Welsh Assembly for South Wales West.
 \\ \hline
\textbf{100-a:} Wales Assembly Member for South Wales West \textcolor{red}{Rachel} Jones says she has not been formally asked to become a UKIP MEP.
 \\ \hline
\textbf{Full supervision:} First Minister \textcolor{red}{Carwyn} Jones has said she is "surprised" she has not been asked to become a UKIP MEP.
 \\ \hline
 \textbf{Gold Summary:} UKIP's Welsh MEP post may be better left unfilled as a result of Brexit , party AM \textcolor{red}{Caroline} Jones has said .
  \\ \hline
\end{tabularx}
\caption{An example of WikiTransfer model output across dataset size used in fine-tuning, illustrating how model output style and hallucinated entities differ as the model moves from Wikipedia pretraining as a source of knowledge to the target dataset. Text not stated in the source document is highlighted in red.}
\label{tab:example_summaries}
\end{table}
}

\newcommand{\TableAllUda}{
\begin{table}[t!]
\tiny
\centering
\resizebox{\columnwidth}{!}{\begin{tabular}{ c | c c c | c c c | c c c | c c c  } \Xhline{2\arrayrulewidth}
             \textbf{Target Dataset} &  \multicolumn{9}{c}{CNNDM}  \\
             \Xhline{2\arrayrulewidth}
             \textbf{Transfer from} & \multicolumn{3}{c|}{WikiTransfer} & \multicolumn{3}{c|}{Reddit} & \multicolumn{3}{c}{BART} \\ \Xhline{2\arrayrulewidth}
              \textbf{10-UDA} & \textbf{39.51} & \textbf{17.2} & \textbf{35.93 } &  38.86 & 16.83 & 35.26  &  37.81 & 16.82 & 34.64 \\
             \Xhline{2\arrayrulewidth}
             \textbf{100-UDA} & 39.89 & 17.27 & 36.26  & \textbf{40.19} & \textbf{17.91} & \textbf{36.98} & 38.43 & 17.22 & 35.49  \\
             \Xhline{2\arrayrulewidth}
             \multicolumn{4}{c}{}   \\
            \Xhline{2\arrayrulewidth} 
            \textbf{Target Dataset} &   \multicolumn{9}{c}{XSum} \\
            \Xhline{2\arrayrulewidth}
            \textbf{Transfer from} & \multicolumn{3}{c|}{WikiTransfer} & \multicolumn{3}{c|}{Reddit} & \multicolumn{3}{c}{BART} \\
            \Xhline{2\arrayrulewidth}
             \textbf{10-UDA}  & \textbf{35.09} & \textbf{12.86} & \textbf{26.94} & 29.34 & 9.56 & 22.6   & 19.75 & 3.19 & 15.09  \\
            \Xhline{2\arrayrulewidth}
            \textbf{100-UDA} & \textbf{36.57} & \textbf{13.89} & \textbf{28.42} & 33.91 & 12.23 & 26.05  & 26.44 & 7.97 & 20.35   \\
            \Xhline{2\arrayrulewidth}
            \multicolumn{4}{c}{}   \\
            \Xhline{2\arrayrulewidth}
            \textbf{Target Dataset} &  \multicolumn{9}{c}{Reddit} \\
            \Xhline{2\arrayrulewidth}
            \textbf{Transfer from} & \multicolumn{3}{c|}{WikiTransfer} & \multicolumn{3}{c|}{CNNDM} & \multicolumn{3}{c}{BART} \\
            \Xhline{2\arrayrulewidth}
             \textbf{10-UDA} & \textbf{27.6} & \textbf{7.19} & \textbf{22.37} & 25.21 & 5.96 & 20.63  &  20.90 & 4.16 & 16.82  \\
            \Xhline{2\arrayrulewidth}
             \textbf{100-UDA} & \textbf{29.91} & \textbf{8.35} & \textbf{23.93 } & 28.28 & 7.68 & 22.83  & 27.67 & 7.46 & 21.81   \\
            \Xhline{2\arrayrulewidth}
             \multicolumn{4}{c}{}   \\
            \Xhline{2\arrayrulewidth}
            \textbf{Target Dataset} &  \multicolumn{9}{c}{BigPatent} \\
            \Xhline{2\arrayrulewidth}
           \textbf{Transfer from} & \multicolumn{3}{c|}{WikiTransfer} & \multicolumn{3}{c|}{CNNDM} & \multicolumn{3}{c}{BART} \\
           \Xhline{2\arrayrulewidth}
              \textbf{10-UDA}  & \textbf{36.58} & \textbf{11.45} & \textbf{32.29} &  33.77 & 9.45 & 29.19   & 32.32 & 10.0 & 28.98  \\  \Xhline{2\arrayrulewidth}
                \textbf{100-UDA} & \textbf{40.25} & \textbf{13.77} & \textbf{35.09 }  &  39.04 & 12.99 & 34.41  & 38.2  & 12.7 & 34.16 5 \\  \Xhline{2\arrayrulewidth}
             
\end{tabular}}
\caption{Results from experiments using the original formulation of UDA \citet{xie2019unsupervised} on 10 examples.}
\label{tab:table_all_uda}
\end{table}
}

\newcommand{\FigureResultsAllBold}{
\begin{figure}[t]
    \centering
    \includegraphics[width=.5\textwidth,height=\textheight,keepaspectratio]{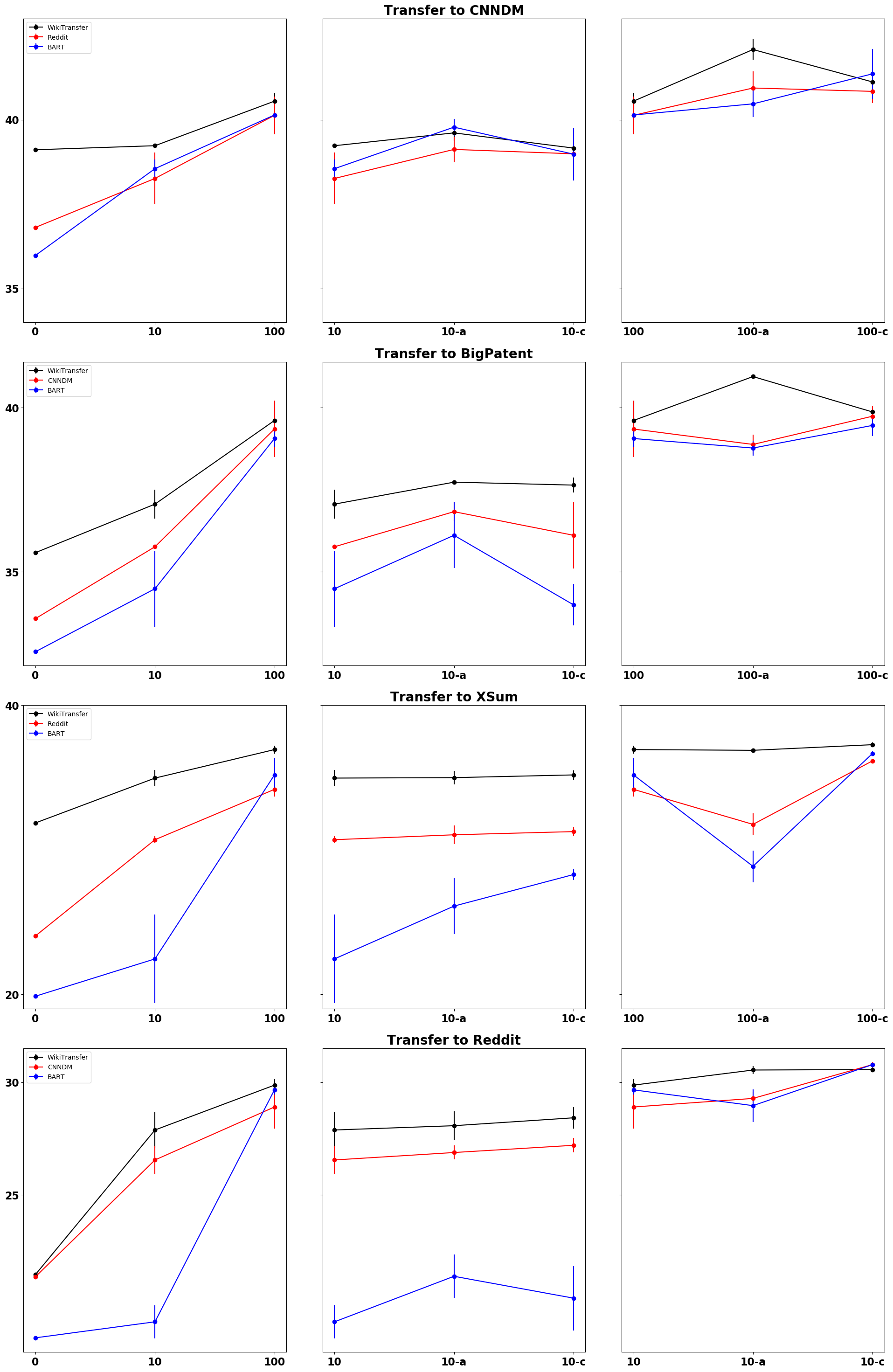}
    \caption{ROUGE-1 scores across datasets, training dataset size, data augmentation (\textbf{*-a}), and consistency loss (\textbf{*-c}) showing the generalizable and robust performance of models transferred from WikiTransfer. Standard deviation bars are also plotted.}
    \label{fig:results_all_plot}
\end{figure} 
}

\title{Improving Zero and Few-Shot Abstractive Summarization \\ with Intermediate Fine-tuning and Data Augmentation}

\author{
 \textbf{Alexander R. Fabbri}\affmark[$\dagger$]
  \quad \textbf{Simeng Han}\affmark[$\dagger\dagger$] \quad \textbf{Haoyuan Li}\affmark[$\ddagger\ddagger$] \\
  \quad \textbf{Haoran Li }\affmark[$\ddagger$] \quad \textbf{Marjan Ghazvininejad}\affmark[$\ddagger$] \quad \textbf{Shafiq Joty}\affmark[$\dagger\dagger$] \\
   \quad \textbf{Dragomir Radev}\affmark[$\dagger$] 
  \quad \textbf{Yashar Mehdad}\affmark[$\ddagger$] \\
\affaddr{\affmark[$\dagger$] Yale University} 
  \affaddr{\affmark[$\ddagger$] Facebook AI} \\
    \affaddr{\affmark[$\dagger\dagger$] Nanyang Technological University}   \affaddr{\affmark[$\ddagger\ddagger$] Renmin University of China} \\
   \email{\{alexander.fabbri,dragomir.radev\}@yale.edu} \\
          \email{\{aimeeli,ghazvini,mehdad\}@fb.com} 
}

\begin{document}
\maketitle

\begin{abstract}
Models pretrained with self-supervised objectives on large text corpora achieve state-of-the-art performance on English text summarization tasks. However, these models are typically fine-tuned on hundreds of thousands of data points, an infeasible requirement when applying summarization to new, niche domains. In this work, we introduce a novel and generalizable method, called WikiTransfer, for fine-tuning pretrained models for summarization in an unsupervised, dataset-specific manner. WikiTransfer fine-tunes pretrained models on pseudo-summaries, produced from generic Wikipedia data, which contain characteristics of the target dataset, such as the length and level of abstraction of the desired summaries. WikiTransfer models achieve state-of-the-art, zero-shot abstractive summarization performance on the CNN-DailyMail dataset and demonstrate the effectiveness of our approach on three additional diverse datasets. These models are more robust to noisy data and also achieve better or comparable few-shot performance using 10 and 100 training examples when compared to few-shot transfer from other summarization datasets. To further boost performance, we employ data augmentation via round-trip translation as well as introduce a regularization term for improved few-shot transfer. To understand the role of dataset aspects in transfer performance and the quality of the resulting output summaries, we further study the effect of the components of our unsupervised fine-tuning data and analyze few-shot performance using both automatic and human evaluation.
\end{abstract}
\section{Introduction}\label{sec:introduction}
Automatic text summarization aims to distill the most salient content of a given text in a compact form. Recent advances in summarization have been driven by the availability of large-scale datasets such as the CNN-DailyMail (CNNDM) corpus \cite{nallapati-etal-2016-abstractive} and the New York Times corpus \cite{sandhaus2008new} as well as by the introduction of large pretrained models such as BART \cite{lewis-etal-2020-bart} and Pegasus \cite{zhang2019pegasus}, in some cases resulting in summaries which are even favored over the human-written reference summaries. 
Creating data for every new domain, however, is infeasible and highly costly. Thus, the ability to transfer large pretrained models to new domains with little or no in-domain data is necessary, especially as such models make their way into real-world applications.  
\par
Unsupervised summarization approaches include autoencoders to mirror the information compression inherent in summarization \cite{baziotis-etal-2019-seq,chu19b,brazinskas-etal-2020-unsupervised} as well as large-scale pretraining for domain-specific adaptation \cite{yang-etal-2020-ted}. However, little work has focused on domain adaptation in summarization. \citet{wang2019exploring} examine domain adaptation for extractive summarization. \citet{hua-wang-2017-pilot} showed that summarization models have difficulty generating text in the style of the target domain, while more recently, \citet{zhang2019pegasus} report strong performance of pretrained models when trained in few-shot settings and \cite{bravzinskas2020few} fine-tune dataset-specific components of a model for few-shot learning. We aim to build on recent work in pretrained models and improve zero-shot and few-shot summarization by encoding characteristics of the target summarization dataset in unsupervised, intermediate fine-tuning data. 
\par
Summarization can be seen as a function of sub-functions of the input, called subaspects, which determine the output form. \citet{jung-etal-2019-earlier} define three subaspects for summarization: position, importance, and diversity, and study how these subaspects manifest themselves in summarization corpora and model outputs. For example, a common subaspect for the CNNDM dataset is position; earlier sentences tend to constitute a good summary.  Inspired by this view of summarization as subaspects, we aim to encode subaspects of a target dataset into unlabeled data to allow a model fine-tuned on this data to learn characteristics of the target dataset to improve zero-shot and few-shot transfer of the model. In our work, we focus on the subaspects of \textit{extractive diversity}, as determined by how well an extractive model performs on the data, \textit{compression ratio} between the source document and summary, and, in the case of CNNDM, the \textit{lead bias}. We assume knowledge of the target dataset such as the size of input documents, the size of the desired summaries, and the extent to which the summary is abstractive, all of that can be treated as prior knowledge if the task is to be well-defined \cite{kryscinski-etal-2020-evaluating}. We encode this knowledge into Wikipedia article data by extracting summaries of the desired output length and filtering examples based on the desired level of abstraction.
\par
Our contributions are the following: 1) We introduce a novel method, called WikiTransfer, which creates pseudo-summaries with subaspects of the target dataset which can be used as unlabeled data for intermediate fine-tuning. We show that this method improves zero-shot domain transfer over transfer from other domains, achieving state-of-the-art unsupervised abstractive summarization performance on the CNNDM dataset while generalizing to other domains, and we perform extensive hyperparameter studies on the factors influencing zero-shot performance 2) We demonstrate the benefits of WikiTransfer in few-shot settings, and show additional improvements when applying WikiTransfer with data augmentation and a regularization term for training with potentially noisy augmented data. We show robustness in these settings and analyze differences in performance in both automatic and human assessments.
\section{Related Work}\label{sec:related_work}
While advances have been made in neural techniques for summarization due in part to large datasets, less work has focused on domain adaptation of such methods in the zero and few-shot settings. \citet{wang2019exploring} examine domain adaptation, but in extractive summarization.  \citet{hua-wang-2017-pilot} examine domain adaptation between opinion and news summarization, observing that models trained on one domain and applied to another domain can capture relevant content but differ in style in generating the summary.
\par 
\citet{bravzinskas2020few} introduce plug-in networks, small finetune-able layers that aim to reproduce characteristics of the target dataset as seen in a small set of labeled examples. In contrast, we aim to encode the characteristics of our target dataset, such as level of extraction and compression, a priori in the intermediate training phase. 
In other work, \citet{lebanoff-etal-2018-adapting} adapt a single-document summarization model to multi-document settings, while \citet{zhu2019transforming} use Wikipedia reference data for downstream query-based summarization
\par
Several approaches for unsupervised summarization have made use of variational autoencoders \cite{baziotis-etal-2019-seq, chu19b, brazinskas-etal-2020-unsupervised}. 
\citet{zhou-rush-2019-simple} makes use of pretrained language models for unsupervised text summarization by aligning the coverage of the generated summary to the source document. \citet{laban-etal-2020-summary} train an unsupervised summarization model with reinforcement learning rewards.  
In another line of work, extractive models such as TextRank, \cite{mihalcea-tarau-2004-textrank}, LexRank \cite{erkan2004lexrank}, and more recently PacSum \cite{zheng-lapata-2019-sentence}, make use of graph centrality for modeling salience.
\par
The power of pretrained models for few-shot transfer was shown for abstractive summarization in \citet{zhang2019pegasus} and extractive summarization in \citet{desai-etal-2020-compressive}. Our work focuses on the zero-shot abstractive summarization setting and the transferability of models fine-tuned on task-specific data from a generic corpus, rather than just the transferability of a single pretrained model. The closest work to ours for zero-shot transfer is \citet{yang-etal-2020-ted}, which uses the lead-bias in news to pretrain an unsupervised model on a large dataset of news articles. Our approach, however, focuses on fine-tuning an already-pretrained model specifically for summarization on a downstream dataset by leveraging a generic text corpus (Wikipedia) to create auxiliary fine-tuning data that transfers across domains, allowing for more fine-grained control over the transfer process. We show the generalizability of such fine-tuning across domains.
BART \citep{lewis-etal-2020-bart} is a pretrained denoising autoencoder and achieved state-of-the-art performance when fine-tuned on summarization tasks at the time. In this work, we use BART as our base pretrained model but in future work will experiment with other pretrained models. 
%
%
%
\section{Methods}\label{sec:methods}
%
%
%
\par \noindent
\textbf{WikiTransfer Intermediate Fine-tuning:}
We propose a method for fine-tuning pretrained models using unsupervised Wikipedia data. We create dataset-specific unsupervised data for this intermediate fine-tuning, by making use of characteristics of the target dataset such as the average length of input documents, the average summary length, and the general bin of whether the summaries desired are very abstractive or very extractive, as discussed above. Assume that we want a summary of M sentences from source documents of N sentences on average, and that we know approximately how extractive the summaries are in the target dataset, as defined as the upper bound ROUGE \citep{lin-2004-rouge} performance of an extractive model, the extractive oracle, on that dataset. We bin the level of extraction of the target summaries into extremely abstractive (ROUGE oracle 10-30), more abstractive (ROUGE oracle 20-30), more extractive (ROUGE oracle 30-50), and extremely extractive (ROUGE oracle 40-60). We then iterate the following procedure on all Wikipedia articles available in a Wikipedia dump: We remove the first M sentences from the Wikipedia article for use as a summary and the following N sentences for use as a source document. Then, we want to check whether this pseudo data point matches the level of extraction of the target dataset. We select the M sentences in the pseudo source document with the highest individual ROUGE scores against the pseudo summary and calculate the ROUGE score between those M sentences concatenated and the pseudo summary, which amounts to a greedy upper bound of the performance of an extractive model on this example. The example will be kept if this ROUGE score falls into the general range of the extractive oracle of the target dataset defined previously and otherwise discarded. We use knowledge of how abstractive a dataset is as a type of summary style which an end-user would know ahead of time. We filter the Wikipedia data points so that only those which fall into the bin for a given dataset are used for fine-tuning. For datasets that are extremely abstractive, such examples may be hard to find, so we remove high-ROUGE sentences from the input until the desired ROUGE oracle score is reached. From here on we refer to data created through this process as \textbf{WikiTransfer}. We then fine-tune a pretrained model on this dataset-specific WikiTransfer data to transfer to a target domain. 
\par \noindent
\textbf{Data Augmentation via Round-Trip Translation:}
In addition to fine-tuning on WikiTransfer data for zero-shot domain transfer, we test the ability of our model to transfer when we have few examples and whether data augmentation further improves these results. In few-shot fine-tuning, we conduct data augmentation to reduce brute-force memorization and introduce a regularization effect. Specifically, we perform round-trip translation \citep{wei2018fast} to generate paraphrases of both the source documents and summaries, as previous work has found this approach creates diverse paraphrase for augmentation while preserving semantic meaning \cite{wei2018fast, xie2019unsupervised}. Our examination found that round-trip translation increased the number of novel n-grams while preserving semantic meaning. Given a dataset of $N$ data points, we translate the source and target sentence-wise into a non-English language and keep the top $k$ beam hypotheses from beam search as output. We then do likewise for the backtranslation to English. This results in $N * k^2$ augmented data points in addition to the $N$ original supervised data points. We align a single beam from the translation to non-English text to a single beam in the backtranslation to English; using all combinations of beams for augmented data did not result in an improvement in initial experiments. We refer to the training setting of $N$ supervised data points with this additional augmented data as $\textbf{N-a}$.
\par \noindent
\textbf{Data Augmentation Consistency:}
While data augmentation may introduce a regularization effect, naively training with augmented data does not necessarily account for noise introduced in the augmented examples. To balance learning from the examples while not overfitting to the small number of supervised samples, the model must learn to be robust to small changes in input examples. We thus investigate the effect of using a consistency loss \cite{xie2019unsupervised, Athiwaratkun2019ThereAM} for few-shot training which enforces consistency between the original and round-trip translated documents with respect to the original summary.
Let $x = \{x_1,x_2,...,x_i,...,x_n\}$ be a source document with $n$ words and $N$ sentences, where $x_i$ represents the $i$-th word in $x$. It could also be represented as $\{s_1,s_2,...,s_j,...,s_N\}$, where $s_t$ represents the $j$-th sentence in $x$. The corresponding target summary $y$ contains $m$ words and $M$ sentences, and $y_i$ denotes the $i$-th token of $y$.
 Standard training, used in the above sections, minimizes the negative log-likelihood loss using supervised teacher forcing \cite{williams1989learning}, which we label  $L_{sup}$:
\begin{equation}
      L_{sup}(x, y) =  -\sum_{t=1}^{m} 
      \log(f(y_t|y_{0:t-1},x,\theta))
     \label{eqn:loss_sup}
\end{equation}

where $f(\cdot|\cdot,\theta)$ represents the distribution among the vocabulary predicted by our model with parameter $\theta$.  
In our formulation, the output (summary) distribution given an augmented (round-trip translated) example should not diverge much from the distribution given the original document, with teacher forcing, so that the model learns to be resilient to small perturbations.
Let $\hat{x}$ be a paraphrase of input document $x$ generated via round-trip translation as described in the previous section. In addition to the supervised loss $ L_{sup}(x,y)$, we introduce another loss $L_{cons}(x,\hat{x},y)$:
\begin{equation}\label{}
\sum_{t=1}^{m}{KL(f(\cdot|y_{0:t-1},x)||f(\cdot|y_{0:t-1,},\hat{x}), \theta))}
\end{equation}
where $KL$ is the KL divergence, which penalizes the model if the probability distribution of the output using the original input is far from the distribution using the round-trip translated input document. Following \citet{xie2019unsupervised}, the gradient does not backpropagate through the model for the distribution of the original input while it does propagate through to the round-trip translated input. 
The total loss $L'$ for training with consistency then is:
\begin{equation}\label{}
L'(x,\hat{x},y)=L_{sup}(x,y) + \lambda L_{cons}(x,\hat{x},y)
\end{equation}
We note that the original formulation of Unsupervised Data Augmentation (UDA) \citep{xie2019unsupervised} enforces consistency in a semi-supervised framework. We also experiment with this setup using unlabeled examples from the target dataset with pseudo labels (for teacher forcing) generated by a model trained on the associated few-shot subset, although this approach is very sensitive to the quality of the pseudo labels (see Appendix). We refer to the training setting of $N$ supervised data points with consistency training as $\textbf{N-c}$.
\section{Experimental Settings}\label{sec:experimental_settings}
\paragraph{Datasets:}
We experiment with four datasets, CNNDM, XSum \cite{narayan-etal-2018-dont}, Reddit\_tifu (Reddit) \cite{kim2018abstractive}, and BigPatent \cite{sharma2019bigpatent}. The datasets were chosen as they all differ in their abstractiveness, output length (from one sentence in XSum to on average four in BigPatent), and cover multiple domains from news (CNNDM and XSum) to social media (Reddit) to patent documents (BigPatent), to show the generalizability of our results. Each of the datasets falls into a different extractive bin, from the most extractive CNNDM to the more abstractive XSum; we discuss these settings further in the Appendix. 
\paragraph{Model Selection and Metric:}
For the experiments which follow, we first choose the model with the best zero-shot performance on a given domain. We test the zero-shot performance from all four domains onto every other domain. For models from our WikiTransfer subset, we choose the best model based on performance on an unsupervised WikiTransfer validation subset. We find that fine-tuning the model longer does not result in performance gains in few-shot transfer, and the checkpoints chosen were typically fine-tuned from 2 to 5 epochs. Results from hyperparameter studies for zero-shot transfer from WikiTransfer data are shown on the validation set of that given target dataset. Unless otherwise stated, all results reported are ROUGE-1/2/L. We run all few-shot transfer experiments on five subsets of supervised data, and the reported numbers, unless zero-shot, are the average of the top three results of the five runs following previous work \cite{gunel2020supervised}. The 10 data point sets are subsets of the 100 data point sets.
\paragraph{Data Augmentation Parameters:}
For data augmentation via round-trip translation, we use a beam size of 10 and $k$ of 10 on German and Russian translation models; fairseq provides bidirectional pretrained translation models \cite{edunov2018backtranslation} from WMT19 \cite{ ng-etal-2019-facebook} for these language pairs. For both 10 and 100 data points, this resulted in 2010 and 20100 total data points. For consistency loss, we use the same augmented data.
%
\paragraph{Model Hyperparameters:}
We use the fairseq codebase \cite{ott-etal-2019-fairseq} for our experiments. Our base abstractive text summarization model is BART-large \cite{lewis-etal-2020-bart}, a pretrained denoising autoencoder with 336M parameters that builds off of the sequence-to-sequence transformer of \citet{vaswani2017attention}. We fine-tune BART using a polynomial decay learning rate scheduler using the Adam optimizer \citep{kingma2014method}.  We mainly vary the learning-rate scheduler, warm-up updates, and total updates. As in the previous few-shot summarization work \cite{zhang2019pegasus} and work in unsupervised machine translation \cite{lample2019cross}, we use a subset of the target-domain validation set for early stopping based on the validation loss. We used the following (warmup updates, total updates, learning rate) parameter tuples based on an examination of the validation curves in initial experiments: 10: (25, 100, 3e-5); 10-a: (20, 200, 3e-5); 100 (20, 200, 3e-5); 100-a: (200, 1000, 1e-5). For consistency loss experiments, we use the $\lambda$ values of 0.1 and 0.5 for experiments with 10 and 100 data points, respectively, chosen manually based on \citet{xie2019unsupervised}. See the Appendix for more details.
\section{Zero-shot Transfer Results}\label{sec:zero_shot_results}
We compare the zero-shot performance of BART fine-tuned on WikiTransfer data to that of one transferred from other summarization datasets. We also show the effect of different choices for WikiTransfer fine-tuning data on CNNDM and XSum.
\subsection{Zero-shot Transfer Comparison}
We aim to show that a model fine-tuned on WikiTransfer data has better zero-shot performance than models transferred from other summarization datasets.
We fine-tune BART on WikiTransfer data for each of the four target datasets described above and also fine-tune a model on each of the fully-supervised datasets. We compare the zero-shot performance of transferring from WikiTransfer against the best zero-shot transfer performance from another dataset in Table \ref{tab:zero_shot}. Zero-shot transfer from WikiTransfer notably outperforms transferring from other datasets on CNNDM, XSum, and BigPatent. On Reddit, we perform better on ROUGE-1 and comparably on ROUGE-2/L, which may be due to distinct writing style on Reddit data, as noted in \citet{zhang2019pegasus}. We also experimented with training a model on data combined from multiple datasets for zero-shot transfer, but this does not report improved results, so for the experiments which follow we use the best performing single-domain transfer model. Details of the fully-supervised BART models are in the Appendix. 
\par
We compare our model to the state-of-the-art unsupervised abstractive model on CNNDM in Table \ref{tab:unsup_comparison}. We outperform the recently-introduced TED model \citep{yang-etal-2020-ted} which was specifically motivated for the news domain. We believe the creation of task-specific data from a generic corpus such as Wikipedia allows for more control over the transfer process than relying on the autoencoder objective of TED, and more generalizable cross-domain results. 
\ZeroShotAllDatasets
\UnsupComparison
\subsection{Effect of WikiTransfer Hyperparameters}
We study the effect the characteristics of our intermediate fine-tuning data have on downstream zero-shot performance on CNNDM and XSum to compare highly extractive and abstractive datasets. 
\par \noindent
\textbf{Effect of learning rate in intermediate fine-tuning:} 
We examine the extent to which overfitting to the unsupervised WikiTransfer data occurs by examining the effect of the learning rate in intermediate fine-tuning on zero-shot transfer performance. We finetune the models on the CNNDM and XSum WikiTransfer data respectively each with a maximum learning rate of 3e-6 and 3e-5. Results are shown in Table \ref{tab:ablations_lr_bin_m}. Using a smaller learning rate in intermediate fine-tuning improves results on CNNDM, but not on XSum, likely due to the simple extractive and lead bias objectives which can easily overfit during fine-tuning. We see a similar trend with the effect of dataset size. For datasets other than CNNDM, we use a learning rate of 3e-5 in intermediate fine-tuning.
\par \noindent
\textbf{Effect of extractive oracle bin use and the choice of M:}
We tested whether using the extractive bin to filter examples in the unsupervised data affected zero-shot transfer. For this experiment, we used the first M sentences from the Wikipedia article as the summary and the remaining N as the source, but do not filter examples according to how extractive they are. From Table \ref{tab:ablations_lr_bin_m}, we see that the extractive bin has a very noticeable effect on transfer results for XSum and a moderate effect on CNNDM. This is to be expected, as the model otherwise is missing information about XSum's distinctive output style. 

We examine how the choice of M affected performance. We set $M=1$ for CNNDM and $M=3$ for XSum and filtered examples in a similar way based on the extractive bin of the target dataset. We see that the choice of M has a large impact on CNNDM performance but no decrease on XSum. This result, combined with the effect of filtering examples based on the extractive bin, gives insight into the importance of the subaspect of abstractiveness over compression for XSum performance.
\par \noindent
\textbf{Effect of intermediate pretraining dataset size:} 
We examined the effect of the size of the WikiTransfer data on downstream performance. Results are shown in Table \ref{tab:dataset_size_comparison}. We see a general increase with the addition of more data, although smaller increases after 100k data points and even a decrease in 250k on XSum, likely due to noise variation. The performance with 10k data points on CNNDM is already much closer to the best performance than the XSum case. We believe that this is due to the highly extractive nature of CNNDM, which is especially easy for a model such as BART to learn, as it is pretrained as a denoising autoencoder. For XSum, we see a noticeable improvement from 10k to 100k examples. We suspect that the abstractive objective is harder for the model to learn with small datasets. As we add more examples, we do not see a noticeable improvement. Such observations agree with our observation of the effect of learning rate and overfitting to the easier CNNDM objective. For the remaining experiments, we use 400k data points based on initial experiments. 
\AblationsLRBinM
\DatasetSizeComparison
\par \noindent
\textbf{Effect of summary sentence choice:}
The first M sentences of a given Wikipedia article were chosen as this introduction intuitively form a coherent summary of the article. We examine the effect of choosing the first sentences compared to choosing based on other criteria. As an alternative, we pick the sentences with the highest self-ROUGE (ROUGE score of a sentence when using all other sentences as the reference summary) in a greedy fashion (the equivalent of the \textbf{IND-ORIG} settings in \citet{zhang2019pegasus}). As in \citet{zhang2019pegasus}, we use ROUGE-1 F1. The sentences chosen under this heuristic consistently corresponded to those which were longest, and the resulting summaries were hence longer. Thus, we also experimented with choosing important sentences by using ROUGE-1 Precision, \textbf{IND-ORIG-P}. The comparison of these methods is shown in Table \ref{tab:sentence_choice_comparison}. The choice of the summary sentence has a noticeable impact on performance. We hypothesize that the coherence lost in the summaries is especially important for the longer CNNDM summaries. Using important sentences other than the first sentence likely adds more diversity in the data, and finding a balance between coherence and output style is an interesting direction for additional work \citep{christensen-etal-2013-towards}. 
\SentenceChoiceComparison
\par \noindent
\textbf{Effect of lead bias on CNNDM fine-tuning:}
We examined the effect of selecting the M sentences greedily chosen for calculating the extractive oracle and inserting them at the beginning of the unsupervised source document versus leaving them in place for CNNDM intermediate fine-tuning. This is meant to mirror the lead bias present in the dataset. This had a slight impact on performance (40.14 vs 39.74 without this bias), and thus we keep the lead bias for CNNDM experiments.
\par \noindent
\textbf{Wikipedia vs target domain unlabeled data:} While Wikipedia is a natural source of unlabeled data, we tested whether creating unsupervised data from unlabeled in-domain data improved results. We performed the same dataset creation treating the source data of the target domain as we did the Wikipedia data. This resulted in about 60k examples for CNNDM and 200k examples for XSum. Fine-tuning on this data, however, resulted in a performance of 38.08/25.83 ROUGE-1 for CNNDM and XSum (vs 39.11/31.85 on WikiTransfer data). The removal of the first sentences may remove too much information in the case of CNNDM, while for XSum, which already has an initial sentence headline removed as the summary, the first sentence may not constitute a very good summary of the remaining document. Wikipedia data often contains multi-paragraph introductions; thus the removal of the first few sentences may still leave a pyramid-structured document with coherent informative content placed at the front. This result supports the emphasis on learning the subaspects of the target domain over simply in-domain training. An analysis of the output of intermediate fine-tuning on CNNDM reveals that the output was more abstractive, due to information present in the summary not being directly stated in the source,  than fine-tuning on Wikipedia. We also experiment with further in-domain pretraining of BART before zero-shot transfer, but this does not result in consistent improvements across datasets. 
\section{Few-Shot Transfer Results}\label{sec:few_shot_results}
We examine whether zero-shot transfer improvements also carry over to the few-shot setting. Also, we explore the effect of data augmentation and consistency regularization techniques. The results of our experiments with varying training data sizes and augmentation methods for all 4 datasets are shown in Figure \ref{fig:results_all_plot} and the Appendix. 
\par \noindent
\paragraph{10 and 100-shot performance with round-trip translation augmentation:}
We see that in few-shot settings, without data augmentation or consistency training, our model outperforms transferring from another domain or vanilla BART. In the case of transfer to Reddit, we observe that despite similar zero-shot performance with transfer from CNNDM, there is a more sizeable gap with 10-shot transfer.  This suggests that our intermediate fine-tuning does more closely align the BART model with the target domain. Furthermore, when training on augmented data from round-trip translation, we see the best performance in transfer from WikiTransfer in all cases except BART transfer to CNNDM on 10-aug, which is likely due to the autoencoder pretraining objective of BART which biases it towards copying and lead bias, allowing it to perform well in applications to CNNDM. We see improvements when training with augmented data in 10-example cases and most 100-example cases for WikiTransfer. Less improvement is seen in the 100-aug setting when transferring from BART or another domain. We hypothesize that the noise present in the larger augmented dataset causes this occasional performance drop, while the WikiTransfer models appear more robust to potential noise. We also found model robustness as the standard deviation of top-performing WikiTransfer models was least among all models in the majority of cases. Interestingly, for transfer from BART and another domain 100-aug only improves on CNNDM, the most extractive dataset, while the largest drop in performance from augmented data occurs on XSum. This XSum performance drop may be caused by the high compression in the XSum summaries which leaves less room for noisy output when compared to the longer CNNDM and BigPatent summaries which may still preserve the main meaning of the original summary better despite backtranslation noise. In most cases, 100-aug with WikiTransfer results in the best performance, only several points from the state-of-the-art supervised performance.
\FigureResultsAllBold
\par \noindent
\paragraph{Transfer with Consistency Training:} 
We find contrasting trends with the added consistency loss compared to data augmentation via round-trip translation. We note the most sizeable improvements in the more abstractive cases of XSum and Reddit. We hypothesize that the consistency loss promotes better abstraction as the model learns to be invariant to noise which does not change the meaning of the text, and is thus equipped with a better notion of paraphrasing. The consistency loss allows for better training of vanilla BART as well as in general better transfer from other domains than without consistency loss. The loss likely provides a regularization factor which prevents the models from overfitting to the supervised examples. As the WikiTransfer model is already more closely tuned to the target domain, this regularization may not make as large of a difference. This aligns with our observation of WikiTransfer models being more robust to noisy backtranslated data on XSum and Reddit. Transfer to Reddit shows similar results across models for consistency loss with 100 examples (better ROUGE-L for WikiTransfer, better ROUGE-1/2 for Reddit); vanilla BART's strong performance at 100 examples suggests that the information provided in this subset is sufficient for good performance, thus diminishing the gains from the head-start the WikiTransfer model provides in zero and 10-shot transfer. We leave aspects of the consistency training such as the role of the quality of the round-trip translation data and its relation to the transfer domain to future work. 
\TableHumanEvaluation
\subsection{Human Quality Assessment}
We examine how the improved performance from WikiTransfer manifests itself in qualitative annotations when varying the amount of training data. We collect human judgment annotations for two of the four quality dimensions studied in \citet{kryscinski-etal-2019-neural, fabbri2020summeval}, namely consistency and relevance. Consistency is defined as the factual alignment between the summary and the summarized source text, while relevance is defined as the selection of important content; only relevant information should be included in the summary. We did not include fluency as a dimension as an initial inspection of the data found fluency to be of very high quality, and we did not include coherence due to our inclusion of single-sentence XSum summaries where coherence is not a factor. We randomly select 50 examples per dataset and collect the model output from the best-performing zero-shot, 10-aug, 100-aug, and fully supervised models on CNNDM and XSum. The annotator sees the source article and randomly-ordered output from the four models rates the summaries for relevance and consistency on a Likert from 1-5, with 5 being the best score. We averaged the score of two native English-speaking annotators on each example and then across examples, and found moderate and strong annotator correlations for relevance and consistency, respectively. Results are shown in Table~ \ref{tab:table_human_evaluation}. For CNNDM, we see an increase in consistency as more training data is added but not a statistically significant difference (using a Student's t-test with a p-value of 0.05) between 100 and full supervision for any of the relevance or consistency results. The relevance of the full model does not outperform the others, likely because the model output was more concise and was judged as not including source information, while the zero-shot output more closely resembles the lead-three bias, so was judged as more informative. For XSum, we see that relevance improves noticeably as more training data is used. We see varied results for consistency, although without statistically significant differences. This fluctuation in scores may be due to the transition of the model from using knowledge from pretraining in its output versus knowledge from the target dataset obtained during fine-tuning, which we discuss in the Appendix. 
\section{Conclusion}\label{sec:conclusion}
We introduced WikiTransfer, a novel and generalizable method for fine-tuning pretrained models on dataset-specific unsupervised data obtained from generic Wikipedia data. WikiTransfer models achieve state-of-the-art zero-shot abstractive summarization performance on the CNN-DailyMail dataset and generalize across three additional datasets. In few-shot settings, WikiTransfer models are robust to noise introduced through data augmentation and benefit from consistency loss on more abstractive datasets. Furthermore, human assessments of the resulting summaries do not show significant differences between the WikiTransfer few-shot summaries and fully-supervised summaries, demonstrating the efficiency of our approach.
\section{Ethical Considerations}\label{sec:ethical_considerations}
We make use of existing datasets available through libraries such as huggingface's datasets library. Biases may exist in the datasets, such as political bias in the news datasets as well as gender bias in potentially all of the datasets. Thus, models trained on these datasets may propagate these biases. When used as intended, applying the summarization models described in this paper can save people much time. However, the current models are still prone to producing hallucinated summaries, and in such a case may contribute to misinformation on the internet. Further research is needed for ensuring the faithfulness of abstractive summaries to address this issue, as this issue is present among all current abstractive summarization models. 
\par
The experiments make use of V100 GPUs. We used up to 8 GPUs per experiment (depending on the experiment; sometimes a single GPU was used to run the maximum number of experiments in parallel). The experiments may take from several minutes in the case of few-shot experiments without augmentation to a couple of hours for the larger augmented datasets, and up to one day for full-dataset training. Over 400 experiments were run due to our requirement of averaging across multiple experiments. Future work should experiment with distilled models for more light-weight training. We note that while our work required extensive experiments to draw sound conclusions, future work will be able to draw on these insights and need not run as many large-scale comparisons, and models in production may be trained once for use using the most promising settings.  
\section{Acknowledgements}
We thank Griffin Adams, Shrey Desai, and the NAACL 
reviewers for their constructive feedback. 
\bibliography{naacl2021}
\bibliographystyle{acl_natbib}
\appendix
\section{Appendix}\label{sec:appendix}
\TableComparePegasus
\ExampleSummaries
\subsection{Comparison to Previous Work} We show a comparison of our best-performing WikiTransfer few-shot results with those from \citet{zhang2019pegasus} in Table \ref{tab:table_compare_pegasus}. The Pegasus numbers were obtained by a single run as opposed to our average of the best three over 5 subsets. We show large improvements with our few-shot approach compared to previous numbers, except for the 100-shot experiment on XSum. The XSum dataset has the highest overlap with the Pegasus pretraining dataset of all datasets explored in \citet{zhang2019pegasus}, although that work states that the effect of removing this overlap does not affect the full-dataset performance. We hope that this comparison promotes future benchmarking of few-shot results.
\subsection{Sample Summary Outputs}
We include an example of model output summaries on the XSum dataset in Table \ref{tab:example_summaries}. The example serves to demonstrate how output style varies as the amount of training data is increased and how the source of pretraining or fine-tuning data affects this style and model hallucinations. The source document does not state the first name of Ms. Jones, yet every model output, and the gold target, give her one. For zero and 10-aug, the model outputs Lorraine Jones, likely still under the influence of BART Wikipedia pretraining, as there is a Wikipedia article on the Welsh politician Ruth Lorraine Jones (although it does not appear in our intermediate fine-tuning subset). The zero and 10-aug also most resemble Wikipedia introduction sentences; although the output is compact and abstractive like an XSum target sentence, the "X is Y" format of Wikipedia appears. We see at 100-aug examples that the model output is stylistically already much like that of the fully-supervised output and gold summary. This stylistic change is also reflected in the change in hallucination; the use of Rachel Jones is likely caused by the appearance of the name of a minister Rachel Haves in an article on Welsh politics found in the 100-aug subset. The model at this point is already fitting strongly to the target domain. For the fully supervised output, we see the use of Carwyn Jones, which does not match the gender of Ms. Jones but which is found 1090 times in the training source documents. Caroline Jones, the actual person in question, only appears 21 times in the training set. This phenomenon points to two interesting research directions for future work, how to properly preserve world knowledge from pretraining and improvement faithfulness to the source text in knowing when to insert world knowledge. 
\TableAll
\subsection{Additional Training Setting Details}
We provide additional details regarding the training and validation of models. We also provide the exact numbers for few-shot transfer in Table \ref{tab:table_all}.
\par \noindent
\textbf{WikiTransfer Data:} 
We use the statistics from the original papers to determine the extractive bin of the dataset except for the case of Reddit; upon seeing the strong zero-shot performance of the CNNDM, we investigated the extractive oracle of the Reddit dataset and found it to be much higher (about 31 ROUGE-1) than that stated in the original paper. We select the first M sentences for the pseudo-summaries from Wikipedia except in the case of Reddit, where we choose the IND-ORIG setting; this did not result in a difference in zero-shot performance but upon a qualitative inspection of the output, we found the IND-ORIG to be less biased towards Wikipedia style with the coherence of the summaries not being an issue. 
\par
We believe that the approximate level of extraction of desired summaries should be treated as prior knowledge. We also examine, however, how many data points are needed to accurately find the extractive oracle bin from target datasets. We found that using 10 data points sufficed to accurately estimate the bin of the extractive oracle.
\par
Using the first M sentences does not produce ideal summaries of the remaining Wikipedia article, but experiments comparing the WikiTransfer approach on Wikipedia data as opposed to using in-domain data, as well as manual inspection of the data showed the validity of using Wikipedia data for proxy summaries. While the extractive oracle provides some measure of overlap, this heuristic does not ensure deeper semantic overlap or faithfulness between the pseudo summary and the rest of the article. We believe a valuable direction for future work is improving the target-specific data as well as encoding additional semantics and style-based subaspects into the pseudo summaries.
\par \noindent
\textbf{Training and Validation Hyperparameters:} We found that full-precision floating-point gave slightly better, and more stable, results, so we report full-precision floating-point numbers. We set a maximum tokens-per-batch of 1024 and use gradient accumulation with an update frequency of 8 for all experiments with 10 data points, and 32 for 10-aug as well as all experiments with 100 (+ augmented) data points. For CNNDM 10 examples, we found it necessary to use a smaller learning rate (3e-6) to avoid immediate overfitting. We perform validation after each model update, as the models typically converge in under 50 iterations. For the 100-aug setting, we begin validation checking after 50 iterations as the models typically converged around 100 iterations.  We train with label-smoothed cross-entropy \cite{szegedy2016rethinking} loss for few-shot transfer. We found that models can be sensitive to the choice of hyperparameters in the few-shot settings, hence the averaging over 5 subsets to reduce variation. 
\par
We use the standard training and testing splits of each dataset (for Reddit, we use the same 80-10-10\% split as in \citet{zhang2019pegasus}), and thus refer the reader to the original papers for detailed statistics. For validation, we used a subset of the target-dataset validation set consisting of 4k examples. While this matches previous unsupervised and transfer settings, we understand that the use of a large validation set is not ideal. We experimented with smaller validation sets on Reddit transfer and found that the results did not change using a validation set of only 10 data points, although we leave a further examination of the effect of validation set size for future work. 
\par
We provide the range of the label-smoothed cross-entropy validation loss by taking the average validation loss (over five subsets) from the best-performing and worst-performing transfer models on a given dataset. The range of validation losses for CNNDM is (4.49, 5.05), for XSum (4.63, 5.45), for Reddit (5.98, 6.65), and for BigPatent (4.88, 6.40). 
\par
\par \noindent
\textbf{Full Supervision and Additional Experiments:} 
For zero and few-shot transfer, we compare transfer from BART trained on WikiTransfer data to the best-transferring BART model trained on the datasets. The following numbers are ROUGE-1. Our application of BART on fully-supervised data achieves state-of-the-art performance on Reddit (32.74). We perform slightly worse on CNNDM (44.16 vs 45.94 from \citet{dou2020gsum}). Lower performance when compared to Pegasus-large \citep{zhang2019pegasus} on XSum (45.14 vs 47.21) and BigPatent (43.34 vs 53.63) is likely due to differences in capacity and training batch size, as our performance is comparable to Pegasus-base. Our approach is not model-specific to BART, so we leave the application of other models such as Pegasus to future work and do not focus on achieving state-of-the-art on the fully-supervised individual datasets.
\par
We limit our primary few-shot experiments to 10 and 100 data points, as we are primarily interested in real-world few-shot applications where we likely do not have 1k data points. Initial experiments using 1k and 10k data points on CNNDM showed that WikiTransfer still outperforms transfer from other domains, although both remain below state-of-the-art performance. We leave a further examination of fine-tuning on larger training sets for future work. 
\subsection{Semi-supervised UDA experiments}
We experimented with the original formulation of UDA in a semi-supervised setting. In this framework, the label (summary) outputted by the model for an augmented example should be the same as the label of the original document on unlabeled examples. Let $x_{U}$ be an unsupervised source document from the target dataset other than our supervised few-shot examples. Let $\hat{x}_{U}$ be a paraphrase of input $x_{U}$ generated via round-trip translation as in our above data augmentation experiments. To apply teacher forcing, we require a label  $y_{U}$, which we obtain for each model by applying the model fine-tuned on the analogous few-shot subset. In addition to the supervised loss $L_{sup}(x,y)$, we thus introduce another loss $L_{uda}(x_{U},\hat{x}_{U},y_{U})=:$
\begin{equation}\label{}
\sum_{t=1}^{m}{KL(f(\cdot|y_{U0:t-1},x_U)||f(\cdot|y_{U0:t-1},\hat{x}_U))}
\end{equation}
\TableAllUda
In practice, for an epoch, we iterate through the supervised examples with loss $L_{sup}$ followed by iterating over the unsupervised examples $L_{uda}$. We sampled 1k unlabeled data points for 10-UDA experiments and 3k unlabeled data points for 100-UDA. Results of initial experiments are shown in Table \ref{tab:table_all_uda}. We find that the performance of the UDA models is very dependent on the quality of the pseudo-labels generated. We chose the model trained on the first data subset of the 5 runs to generate the pseudo-labels and if this model had higher performance then this model likely performed better in UDA (this occurred in our Reddit transfer to CNNDM with 100 data points. As a result, as the quality of the pseudo-labels improves with 100-shot training the UDA performance improves and is more comparable to the unaugmented performance in Table~\ref{tab:table_all}.






\end{document}